\documentclass[final]{article}

 \usepackage[creativeai]{neurips_2025}

\usepackage[utf8]{inputenc} 
\usepackage[T1]{fontenc}    
\usepackage{hyperref}       
\usepackage{url}            
\usepackage{booktabs}       
\usepackage{amsfonts}       
\usepackage{nicefrac}       
\usepackage{microtype}      
\usepackage{xcolor}         
\usepackage{graphicx}

\title{Real-Time Intuitive AI Drawing System for Collaboration: Enhancing Human Creativity through Formal and Contextual Intent Integration}

\author{
  Jookyung Song\textsuperscript{1} \\
  \texttt{chsjk9005@gmail.com}
  \And
  MooKyoung Kang\textsuperscript{1,2} \\
  \texttt{nostone@gmail.com}
  \And
  Nojun Kwak\textsuperscript{1} \\
  \texttt{nojunk@snu.ac.kr}
}

\begin{document}

\maketitle

\begin{center}
\textsuperscript{1}Seoul National University \\
\textsuperscript{2}Xorbis Co., Ltd.
\end{center}

\begin{abstract}
  This paper presents a real-time generative drawing system that interprets and integrates both \textbf{formal intent}—the structural, compositional, and stylistic attributes of a sketch—and \textbf{contextual intent}—the semantic and thematic meaning inferred from its visual content - into a unified transformation process. Unlike conventional text-prompt-based generative systems, which primarily capture high-level contextual descriptions, our approach simultaneously analyzes ground-level intuitive geometric features such as line trajectories, proportions, and spatial arrangement, and high-level semantic cues extracted via vision–language models. These dual intent signals are jointly conditioned in a multi-stage generation pipeline that combines contour-preserving structural control with style and content-aware image synthesis. Implemented with a touchscreen-based interface and distributed inference architecture, the system achieves low-latency, two-stage transformation while supporting multi-user collaboration on shared canvases. The resulting platform enables participants, regardless of artistic expertise, to engage in synchronous, co-authored visual creation, redefining human–AI interaction as a process of co-creation and mutual enhancement.

\end{abstract}

\section{Introduction}

The rapid evolution of artificial intelligence (AI) has transformed artistic creation from a tool-driven practice into a mode of human–machine co-creativity. From early generative adversarial networks\citep{goodfellow2020generative} to diffusion-based and real-time generation models, AI has shifted from assisting isolated creative tasks to participating in continuous, collaborative processes. This transition has expanded the scope of human imagination, yet it has also revealed fundamental gaps in how AI interprets human creative intent.

Most contemporary generative systems rely on \textit{text prompts}, which primarily capture a creator's \textbf{contextual intent}—the semantic and thematic description of what is to be generated. While effective for language-driven ideation, this paradigm overlooks non-verbal, sensory modes of expression such as drawing. As prior work has noted \citep{mccormack2019autonomy}, the predominance of textual input constrains the representation of visual thinking, embodied gestures, and spatial reasoning.

Recent research has attempted to incorporate form-based inputs into generative pipelines, yet these systems often fail to preserve a user's \textbf{formal intent}—the structural attributes of a sketch such as line quality, proportion, spatial composition, and rhythm—particularly in spontaneous or casual drawing contexts. This misalignment can reduce the sense of creative agency, relegating AI to the role of an after-the-fact editor rather than an equal creative partner \citep{mammen2024creativity}. Such limitations have further fueled debates around authorship and creative responsibility in AI-enhanced art \citep{harvardlawreview2025doublebind}, where the boundaries between human and machine contribution remain contested.

These challenges motivate the central question of this work: \textit{How can a generative system simultaneously interpret and integrate both formal and contextual intent in real time, while enabling meaningful AI augmentation?} Our system addresses this challenge by capturing user sketches through a multi-touch interface, extracting structural features with contour and edge detection, and combining them with semantic cues derived from a vision–language model. To ensure precise preservation of \textbf{formal intent}, we apply a Concave Hull–based masking method that tightly conforms to the user’s drawn silhouette. This approach not only prevents unintended spillover during generation, but also allows fine adjustment of stroke thickness, enhancing the perception of shape and depth in the generated output. By accurately defining the mask boundary, the system enables seamless pixel-level stitching with surrounding regions, ensuring that AI-generated content integrates smoothly without degrading the original hand-drawn qualities. This structural conditioning is further reinforced through ControlNet, while contextual intent guides thematic composition and background synthesis. For background integration, we employ LoRA models fine-tuned on specific background styles, allowing the generated content to blend naturally with the inferred scene context. A two-stage generation process with an distilled diffusion model produces results in under two seconds, enabling multi-user, synchronous co-creation.

Beyond its technical contributions, the system has demonstrated notable sociocultural impact by fostering collective creativity, lowering barriers to artistic participation, and promoting embodied forms of expression. Deployed as \textbf{Graffiti-X} in public installations, it has engaged diverse audiences in collaborative art-making, generating significant public interest and reinforcing the role of AI as an inclusive creative partner.

\begin{figure}[t]
    \centering
    \includegraphics[width=\linewidth]{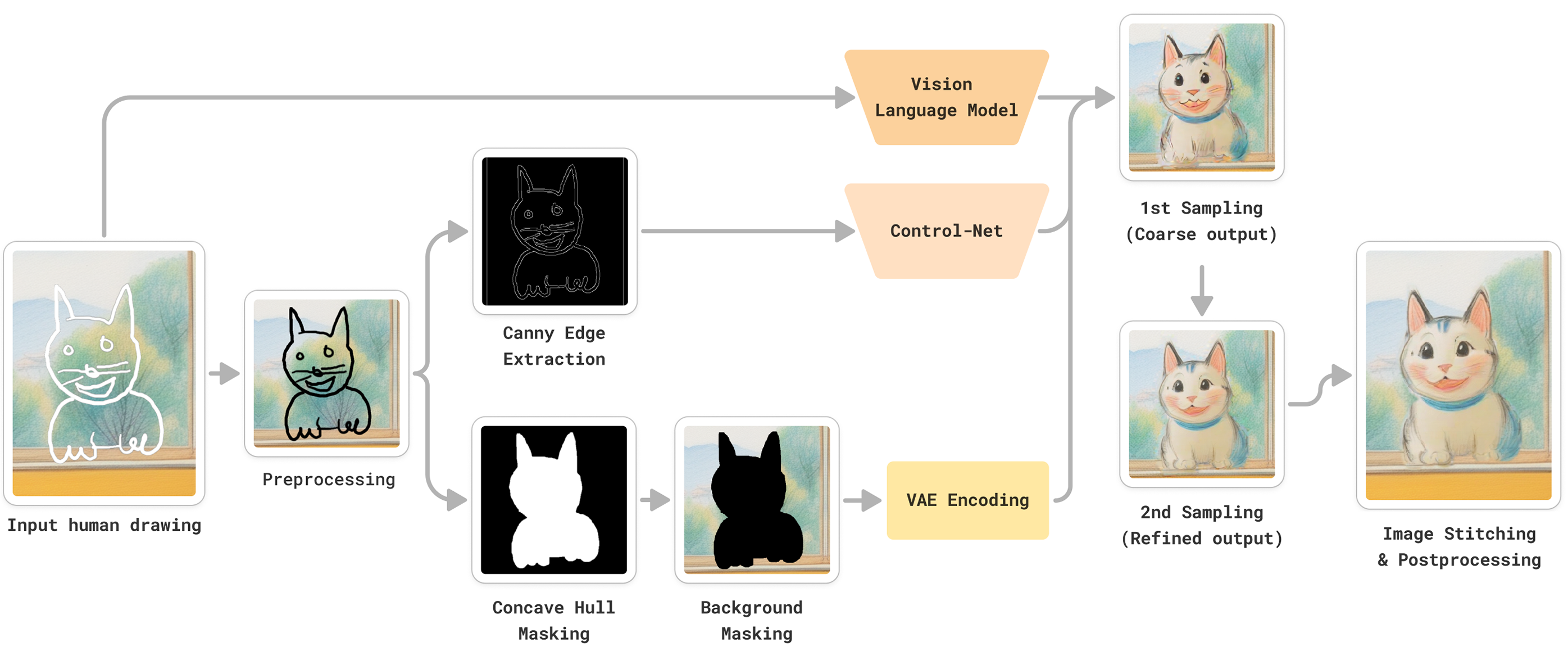}
    \caption{
        Overview of the proposed real-time image drawing pipeline, from input to AI-guided generation and final compositing.
    }
    \label{fig:pipeline}
\end{figure}
\section{Related Works}

Recent studies have explored diverse approaches to human–AI co-creativity, offering important insights for the design of our system. Lin et al. \citep{lin2023beyond} investigate mixed-initiative co-creativity systems that allow humans and AI to share creative control, informing our real-time feedback loop and collaborative composition strategies. Ibarrola, Lawton and Grace \citep{ibarrola2023collaborative} present CICADA, a context-aware drawing agent capable of inferring semantic meaning from sketches to complete vector-based illustrations, demonstrating the value of integrating formal intent. Zhao et al. \citep{zhao2025reframer} introduce Reframer-HRC, which enables multi-user collaborative drawing with robotic and AI interaction, highlighting the feasibility of synchronous co-creative environments. Similarly, Lin, Agarwal, and Riedl \citep{lin2022creative} propose Creative Wand, a framework for human-AI communication in co-creation, emphasizing the importance of real-time responsiveness and intent interpretation. 

In a related domain, \citet{song2024sketcherx} explores portrait-based human–robot co-creation. It was recognized as an interactive art project that investigates the performative and relational aspects of human–robot creative interaction. Its integration of visual transformation, machine interpretation, and embodied robotic execution illustrates how AI-driven creative systems can bridge digital and physical media while engaging audiences in novel artistic experiences.
Together, these works underscore the transition from prompt-based systems toward real-time, form-driven, collaborative generative platforms, a trajectory that directly motivates the approach presented in this paper.

\section{System Overview}

The proposed system is a real-time creative platform that allows multiple users to simultaneously draw on a shared large touchscreen, and have their sketches transformed into coherent, stylized digital artworks. The transformation process is guided by both \textit{Formal Intent}---the structural and compositional characteristics of the drawing---and \textit{Contextual Intent}---the semantic meaning, mood, and thematic cues. To integrate both intents, the system combines contour and edge-based structural preservation with vision–language–driven semantic guidance in a unified generation pipeline.

\subsection{Image Processing Pipeline}
\begin{figure}[t]
    \centering
    \includegraphics[width=\linewidth]{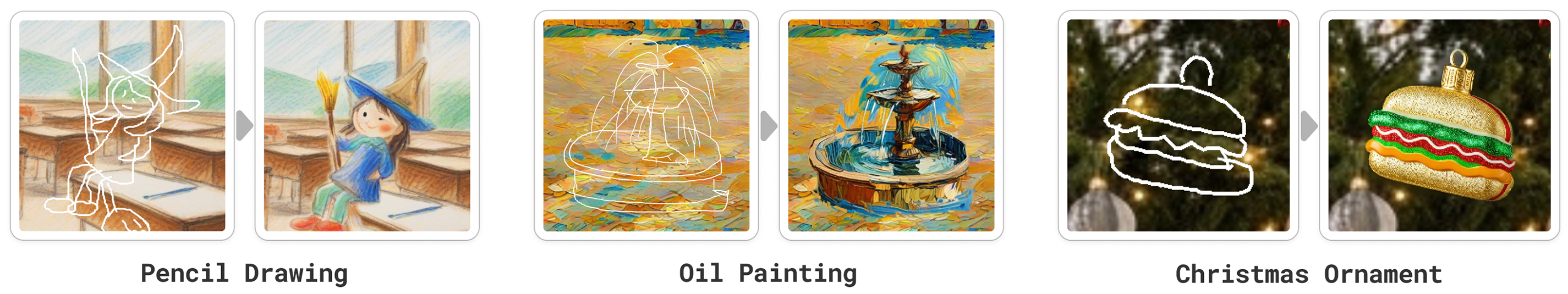}
    \caption{
        Style transformation results from the proposed system. 
        Appropriate LoRA modules, selected based on extracted \textit{Contextual Intent}, adapt the background style (e.g., pencil drawing, oil painting, Christmas ornament) while ControlNet-based conditioning and Concave Hull masking preserve the user’s \textit{Formal Intent} in line quality, proportion, and composition. 
        The generated content blends seamlessly into the background through Contextual Style-Aware Inpainting.
    }
    \label{fig:style_examples}
\end{figure}

\textbf{Dual Intent Integration.} 
The proposed system implements a dual-path processing pipeline to integrate both \textit{Formal Intent} and \textit{Contextual Intent} in real time. 
Formal Intent—structural and compositional properties of the sketch—is preserved through Concave Hull-based masking and ControlNet-driven structural conditioning, while Contextual Intent—semantic, thematic, and affective cues—is extracted via a vision–language model (VLM) and converted into generation prompts.

\textbf{(1) Multi-touch capture and pre-processing.} 
User strokes are captured in real time using TouchDesigner, which assigns a persistent ID to each contact point and streams the curves to the processing pipeline. 
The system supports heterogeneous devices and automatically normalizes aspect ratios when necessary, using pixel expansion techniques to avoid distortion.

\textbf{(2) Masking and structural extraction (Formal Intent).} 
The sketch region is isolated using a \textit{Concave Hull} algorithm, which tightly conforms to the user’s drawn silhouette, preserving contour thickness and overall shape. 
This precise masking prevents unintended spillover during generation and facilitates seamless blending with surrounding pixels. 
Canny edge detection is applied to produce high-fidelity edge maps, ensuring that ControlNet receives detailed structural information.

\textbf{(3) Contextual intent extraction.} 
A CLIP-based vision–language model analyzes the masked sketch and surrounding context to extract semantic descriptors, thematic keywords, and emotional tone. 
These are automatically converted into text prompts that encapsulate the inferred Contextual Intent.

\textbf{(4) Latent-space encoding and inpainting setup.} 
The masked image and corresponding edge map are encoded into a latent space representation via a variational autoencoder (VAE). 
Contextual Style-Aware Inpainting ensures generated regions blend seamlessly with the existing background, while pixel blending smooths mask boundaries to avoid visible seams.

\textbf{(5) Conditional generation (Formal Intent preservation).} 
ControlNet-based conditioning enforces alignment between the generated output and the original stroke structure. 
A two-stage generation process using KSampler—coarse pass for layout and lighting, followed by a refined pass for detail enhancement—is used. 
The denoising rate is set to 0.3, balancing structure preservation with the introduction of novel detail.

\textbf{(6) Style adaptation (Contextual Intent reflection).} 
Based on the contextual prompt, appropriate LoRA modules are applied to adapt the background style and ensure visual coherence between generated objects and the surrounding scene. 
As shown in Figure~\ref{fig:style_examples}, the system supports diverse styles (e.g., pencil drawing, oil painting, Christmas ornament) that integrate naturally while retaining both formal and contextual Intent.

\textbf{(7) Post-processing and compositing.} 
The generated output is restored to the original canvas size, with temporary pixel padding removed. 
Seam-aware stitching merges the AI-generated imagery with the original strokes, maintaining stroke sharpness and overall visual harmony.

\begin{figure}[t]
    \centering
    \includegraphics[width=\linewidth]{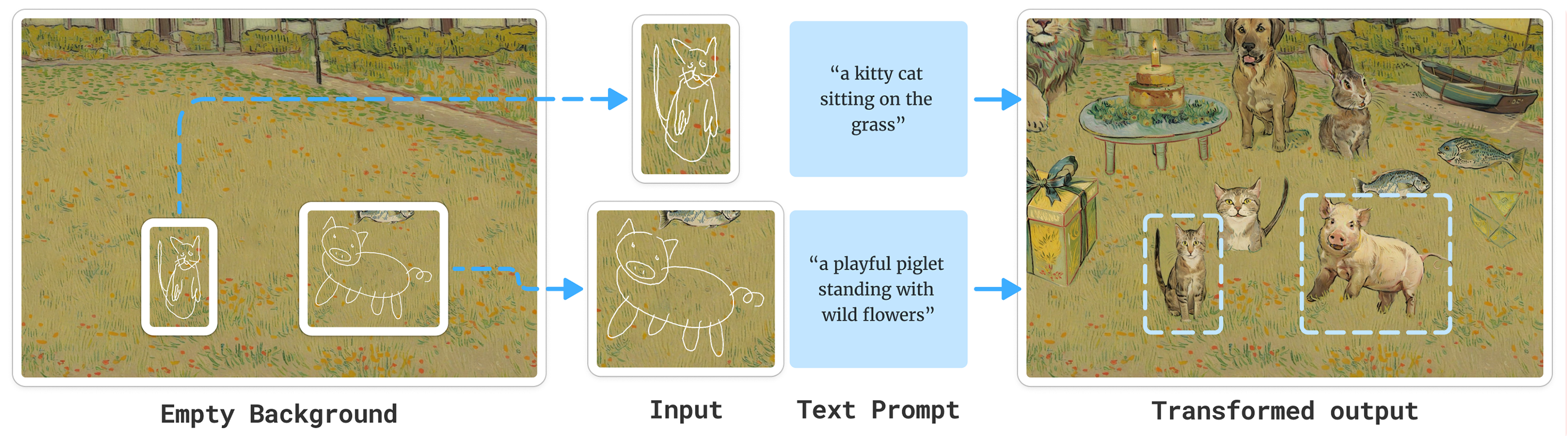}
    \caption{
        Examples of user input and ai transformation. On the left, there is empty background and on the right, there is text prompt which is generated by the vision-language model. 
    }
    \label{fig:input_transformation}
\end{figure}

\subsection{Distributed Inference Architecture}

The system is implemented as a distributed, event-driven architecture optimized for collaborative use. TouchDesigner clients communicate with a central queue server via WebSockets, enabling low-latency transfer of stroke data and AI-generated results. Multiple GPU workers process jobs in parallel, each handling intent extraction, latent-space encoding, and conditional generation. To minimize contention, the shared canvas is spatially partitioned into tiles, allowing concurrent updates in separate regions.

The interaction management module handles blob merging for overlapping strokes, lifetime control to remove stale elements, and resource optimization to prevent performance degradation during extended sessions. Telemetry (queue depth, GPU utilization, per-stage latency) is continuously monitored, and priority scheduling is applied to interactive tasks to maintain median end-to-end latency near two seconds even under peak load.

This integrated design enables an immersive, synchronous co-creation experience in which AI serves as an active collaborator, augmenting human expression in real time while respecting the original creative intent.

\section{Application: Graffiti-X}

\begin{figure}[t]
    \centering
    \includegraphics[width=\linewidth]{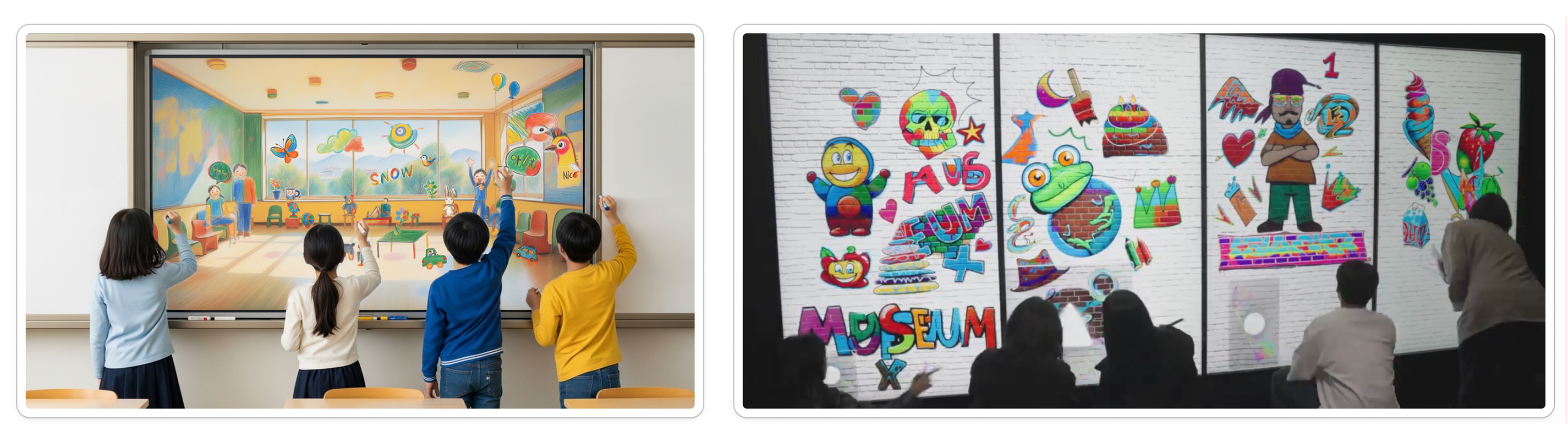}
    \caption{
        Examples of Graffiti-X in use: (Left) collaborative drawing by young students in an educational setting; (Right) large-scale public installation at \textit{Museum X} in Sokcho, South Korea.
    }
    \label{fig:graffiti_examples}
\end{figure}

\textbf{Graffiti-X} is a collaborative drawing platform in which multiple users create sketches on a shared touchscreen, with the AI system interpreting both \textit{Formal Intent} and \textit{Contextual Intent} to generate a unified artwork in real time. The platform supports diverse use cases, including collaborative art education, interactive exhibitions, and community-based creative projects, enabling visual dialogue and participatory creation among users (Figure~\ref{fig:graffiti_examples}).

Beyond artistic contexts, the system can be extended for commercial and educational purposes. In advertising or event-based installations, it can produce audience-driven visuals that adapt to brand themes in real time. In classrooms, it can serve as a hands-on tool for teaching composition, storytelling, and creative collaboration, allowing students to explore visual communication collectively. This versatility positions \textbf{Graffiti-X} as both an artistic medium and a flexible content creation platform for cultural, commercial, and educational settings.
\footnote{Graffiti-X has been deployed as a permanent installation at Museum X in Sokcho, Republic of Korea, and featured as a participatory media art experience at the 2025 K-ICT event in BEXCO, Busan, South Korea.}


\section{Sociocultural Impact}

The proposed system offers a new cultural paradigm for creative collaboration in the era of generative AI, in which human intuition and machine augmentation operate as intertwined agents:

\begin{itemize}
    \item \textbf{Redefining creative agency.} In this system, the act of drawing is interpreted by AI and transformed into an artwork, shifting authorship from a single fixed entity to a shared domain between human and machine. This aligns with prior discussions on co-creative agency \citep{mccormack2019autonomy, loivaranta2025spectrum}, where creative responsibility and authorship are fluid and negotiated in real time.
    
    \item \textbf{Democratization of art.} By enabling non-experts to produce high-quality results through an intuitive, gesture-based interface, the system lowers barriers to participation and expands access to artistic creation, illustrating how AI tools can foster inclusivity in creative practice.
    
    \item \textbf{Fostering collective creation in public spaces.} Multi-user, real-time collaboration, as exemplified by \textbf{Graffiti-X}, facilitates the emergence of \emph{collective art rituals} in community spaces, educational institutions, and exhibitions, strengthening social bonds through shared visual dialogue.
    
    \item \textbf{Restoring embodied creativity.} Unlike text-prompt-based generative tools, this system centers on the embodied act of drawing and the visual language of composition, allowing creators’ sensory and spatial intentions to be directly embedded in the generated outcomes.
    
    \item \textbf{Addressing the ``creative double bind.''} As noted by \citet{harvardlawreview2025doublebind}, human–AI co-creation often places the human in a tension between being the originator of an idea and relinquishing control over the output. By structuring the generation process as a response to human expressive acts, the system offers a framework to mitigate this tension while preserving both \emph{Formal} and \emph{Contextual} Intent.
\end{itemize}

This integration of embodied human creativity with responsive AI generation has the potential to reshape not only how art is produced, but also how authorship, collaboration, and cultural participation are understood in the context of emerging creative technologies.

\section{Conclusion}       

We have presented a real-time generative drawing system that jointly interprets and integrates both \textit{formal intent} and \textit{contextual intent} into a unified co-creative process. By combining structural preservation via contour- and edge-based conditioning with semantic enrichment from vision–language models, our approach enables AI to act not as a post-processing filter, but as an active, synchronous collaborator. The system’s distributed inference architecture supports multi-user interaction with sub-two-second latency, making it suitable for public installations, educational environments, and commercial applications.

Through deployments such as Graffiti-X, we have demonstrated that the proposed framework can democratize high-quality artistic production, foster collective creativity, and reframe human–AI interaction as shared authorship. The integration of embodied input with contextual reasoning allows creators—regardless of skill level—to produce visually coherent and thematically rich works in real time.

Future work will extend the notion of intent beyond the sketch itself, incorporating \textit{contextual intent} derived from the social, spatial, and temporal context in which creation occurs. We also aim to explore adaptive models that learn from repeated human–AI interactions, enabling evolving co-creative relationships. Ultimately, we envision this system as a foundation for a broader class of AI tools that augment, rather than replace, human creativity—shaping not only the artifacts produced but also the cultural practices of making.

\section*{Acknowledgements}
The authors would like to thank Minsung Jung for developing real-time rendering and multi-user interaction modules. This project was conducted as part of an internal R\&D initiative at XORBIS Co., Ltd.

{
    \small
    \bibliographystyle{neurips}
    \bibliography{references}
}

\end{document}